\documentclass[letterpaper, 10 pt, conference]{ieeeconf}  

\IEEEoverridecommandlockouts                              

\usepackage{amsmath}
\usepackage{xspace}

\usepackage[utf8]{inputenc} 
\usepackage[T1]{fontenc}    
\usepackage{hyperref}       
\usepackage{url}            
\usepackage{booktabs}       
\usepackage{amsfonts}       
\usepackage{nicefrac}       
\usepackage{microtype}      
\usepackage{graphicx}
\usepackage{subfigure}
\usepackage{amsthm}
\usepackage{amssymb}
\usepackage{wrapfig}
\usepackage{dsfont}
\usepackage[ruled, noend]{algorithm2e}
\usepackage{mathrsfs}
\usepackage{multicol}
\usepackage{threeparttable}
\usepackage{multirow}
\usepackage{comment}
\usepackage{bm}
\usepackage{pifont}
\usepackage{threeparttable}
\usepackage{dsfont}
\usepackage{makecell}
\usepackage[dvipsnames, table]{xcolor}
\usepackage{caption}
\usepackage[export]{adjustbox}
\usepackage{tcolorbox}
\usepackage{cite}

\newcommand{\qa}[1]{{\textcolor{RoyalBlue}{{#1}}}}

\title{\LARGE \bf
Generalize by Touching: \\ Tactile Ensemble Skill Transfer for Robotic Furniture Assembly
}

\author{Haohong Lin$^{1,2}$, Radu Corcodel$^{2}$ and Ding Zhao$^{1}$ 
\thanks{*This work was fully supported by Mitsubishi Electric Research Labs (MERL)}
\thanks{$^{1}$Department of Mechanical Engineering, Carnegie Mellon University, Pittsburgh, PA 15213 USA, {\tt\small \{haohongl, dingzhao\}@cmu.edu}}%
\thanks{$^{2}$Mitsubishi Electric Research Labs (MERL), Cambridge, MA 02139 USA, {\tt\small corcodel@merl.com}} %
}
\date{June 2023}

\begin{document}

\newtcbox{\mybox}[1][red]
  {on line, arc = 0pt, outer arc = 0pt,
    colback = #1!10!white, colframe = #1!50!black,
    boxsep = 0pt, left = 1pt, right = 1pt, top = 2pt, bottom = 2pt,
    boxrule = 0pt, bottomrule = 1pt, toprule = 1pt}
    
\maketitle 
\thispagestyle{empty}
\pagestyle{empty}

\begin{abstract}

Furniture assembly remains an unsolved problem in robotic manipulation due to its long task horizon and nongeneralizable operations plan. This paper presents the Tactile Ensemble Skill Transfer (TEST) framework, a pioneering offline reinforcement learning (RL) approach that incorporates tactile feedback in the control loop. TEST's core design is to learn a skill transition model for high-level planning, along with a set of adaptive intra-skill goal-reaching policies. Such design aims to solve the robotic furniture assembly problem in a more generalizable way, facilitating seamless chaining of skills for this long-horizon task. We first sample demonstration from a set of heuristic policies and trajectories consisting of a set of randomized sub-skill segments, enabling the acquisition of rich robot trajectories that capture skill stages, robot states, visual indicators, and crucially, tactile signals. Leveraging these trajectories, our offline RL method discerns skill termination conditions and coordinates skill transitions. Our evaluations highlight the proficiency of TEST on the in-distribution furniture assemblies, its adaptability to unseen furniture configurations, and its robustness against visual disturbances. Ablation studies further accentuate the pivotal role of two algorithmic components: the skill transition model and tactile ensemble policies. Results indicate that TEST can achieve a success rate of 90\% and is over 4 times more efficient than the heuristic policy in both in-distribution and generalization settings, suggesting a scalable skill transfer approach for contact-rich manipulation. 

\end{abstract}

\section{Introduction}



Robotic furniture assembly is regarded as one of the most complex problems within the field of robotic manipulations given its contact-rich, long-horizon nature~\cite{todorov2012mujoco,zhu2020robosuite,makoviychuk2021isaac,lee2021ikea,heo2023furniturebench, yuan2017gelsight, lin20239dtact}. The contextual purpose of the objects and the associated sub-tasks that must be executed to succeed the overall task. 
\begin{figure}[h]
    \centering
    \includegraphics[width=0.87\columnwidth]{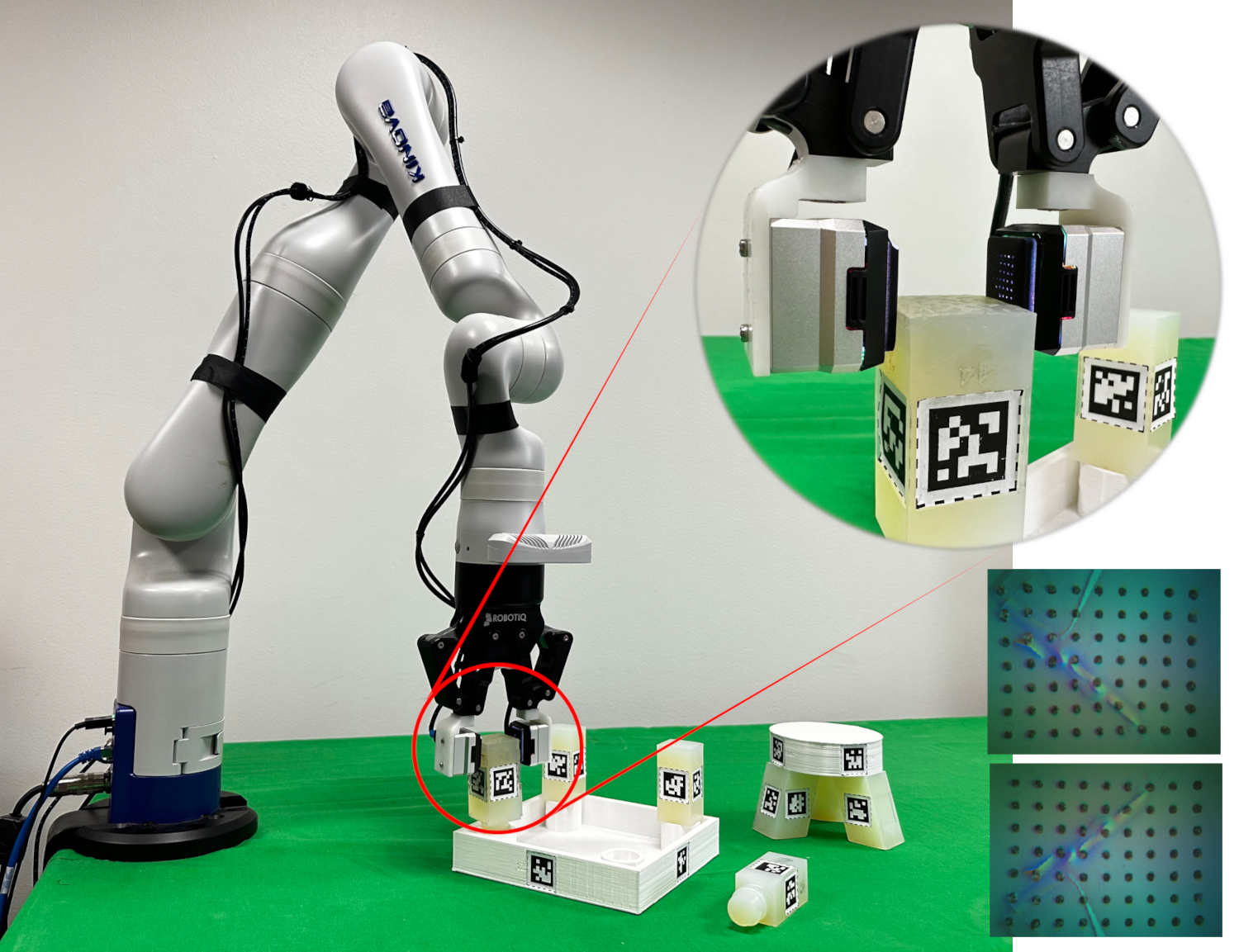}
    \caption{Furniture assembly robot. A natural pipeline of such assembly tasks requires pick, reach, insert, adjust, and thread as the candidate skills to be learned.}
    \label{fig:front_figure}
\end{figure}

\begin{figure*}[h]
    \centering
    \includegraphics[width=1.0\textwidth]{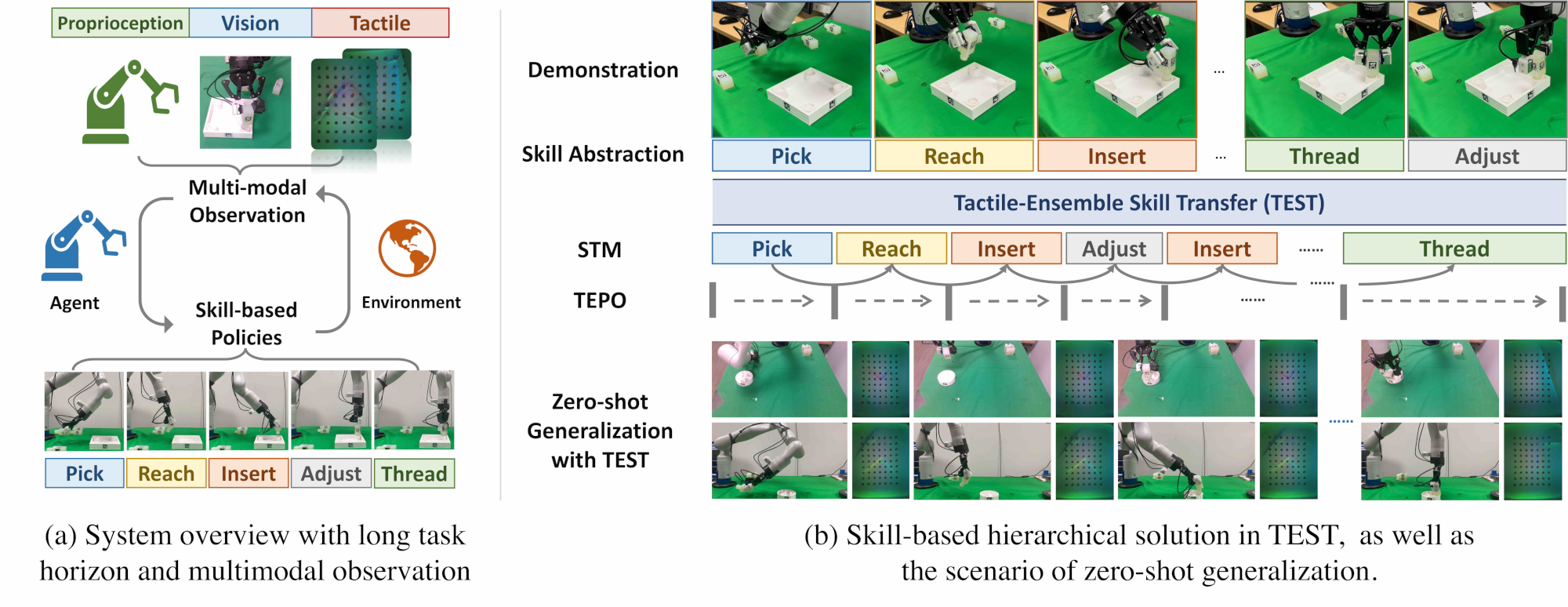}
    \caption{Motivation and challenges in contact-rich robotic assembly problem. }
    \label{fig:motivation}
\end{figure*}

Figure~\ref{fig:front_figure} shows a typical real-world scenario where a robot is tasked to assemble two geometrically distinct by \textit{functionally} identical objects: a three-legged and a four-legged table. The global tasks require the same set of robot skills: picking, insertion, and threading. A common way of assembling these skills in a working robotic platform is by Learning from Demonstration (LfD).
LfD allows robots to learn a policy from humans or heuristic demonstrations. In real-world applications however, LfD is challenging due to its long task horizon and the multimodal nature of the observations, as shown in Figure~\ref{fig:motivation}(a).

The primary concern arises from the \textbf{multimodal inputs} that robots must rely on to observe their environment. With various sensor modalities, there's an inherent uncertainty in the provided data because not all modalities carry meaningful information at the same time during the task. The question then becomes: How can one ensemble multimodal inputs and make accurate predictions without the need for online interactions? 
The challenges do not end with sensor uncertainty. Robotic assembly tasks are implicitly \textbf{long-horizon} in nature. This means that robots need to plan, execute, and connect a series of relevant actions over an extended period of time to achieve the desired global outcome. Traditional LfD approaches, such as Behavioral Cloning (BC), often fall short in these scenarios. They lack the high-level skill awareness required for such complex tasks and struggle with generalization, especially when there is a change in certain sub-task modules in the deployment stage. 

Given the aforementioned challenges, two primary solutions have been proposed. The first is to use ensemble policies that leverage self-supervised learning to counteract the uncertainties from multimodal sensors~\cite{lee2019making, wang2023mimicplay, xu2023xskill}. The second is to use hierarchical Reinforcement Learning (RL) methods to abstract and simplify long-horizon tasks~\cite{daniel2012hierarchical, akrour2018regularizing, hou2020data, pertsch2021accelerating}.

In this paper, we introduce a new approach in Figure~\ref{fig:motivation}(b) that addresses both challenges simultaneously. We present the Tactile Ensemble Skill Transfer (TEST) framework, a unified solution to jointly model the human preferences~(reward), multimodal observations, and actions of robotic assembly tasks.
Our contributions to this work are listed below:

\begin{itemize} 
    \item We formulate robotic assembly as a skill-based RL problem over Goal-conditioned Partially Observable Markov Decision Process~(GC-POMDP, in Section~\ref{sec:objective}) that describes the goal-reaching problem under multimodal sensor inputs instead of the fully observable settings. 
    \item Building on this foundation, we introduce \textbf{T}actile \textbf{E}nsemble \textbf{S}kill \textbf{T}ransfer~(TEST), a skill-based offline RL method that seamlessly integrates the strengths of ensemble learning with tactile feedback and skill-conditioned policy learning.
    \item We validate our approach with real-world experiments using the Furniture Bench platform~\cite{heo2023furniturebench}, evaluating the accuracy and efficiency of learned policy. We also empirically study the generalizability of TEST towards unseen furniture assemblies, and consistency under visual disturbances, highlighting the significant improvements that characterized our framework.
\end{itemize}


\section{Related Work}

\textbf{Contact-rich Robotic Manipulation} 
Our furniture assembly problem originates in the field of contact-rich robotic manipulation. Computer Vision for robotic systems, while pivotal in parsing the semantic understanding of environments, cannot deliver robust information for contact-aware sensing needed to fully close the loop on intelligent robot assembly. This led to the integration of force/torque sensors and later, artificial tactile sensors~\cite{luo2017robotic, lee2019making, luo2021vitac, dong2021tactile} which are crucial in robotic assembly tasks~\cite{hou2020data, nottensteiner2021towards, belousov2022robotic}. We highlight two particular methods of feedback control for robotic assembly.
The first is an end-to-end deep reinforcement learning design, which integrates multiple sensor inputs into a unified framework, allowing for a more holistic understanding of the environment and the task at hand~\cite{akkaya2019solving, chen2022visuo}.
The second is a hierarchical design with skill primitives plus task planning, where each task is broken down into a hierarchy of skills~\cite{akrour2018regularizing, hou2020data, nottensteiner2021towards, belousov2022robotic} managed by specific scripted or learned controllers. This method gained more popularity in robotics research because it allowed for more generalizable and modular solutions. 

\textbf{Skill-based Reinforcement Learning}
The challenges of generalization and long-horizon tasks in robot learning led to hierarchical skill-based policies, often termed as \textit{options}~\cite{sutton1999between, konidaris2009skill, pertsch2021accelerating}. These policies, comprising a high-level skill planner and a low-level controller, aimed to reduce sample complexity and enhance interpretability~\cite{sun2023interactive}. This framework proved especially beneficial for multi-modal decision-making in robotic manipulation~\cite{daniel2012hierarchical, akrour2018regularizing, hou2020data, pertsch2021accelerating}. Skill-based RL's primary challenges are skill discovery and skill chaining. Earlier works either utilized manually designed skill graphs~\cite{NIPS2009_e0cf1f47} or employed unsupervised clustering-based methods~\cite{akrour2018regularizing, srinivas2016option}. However, clustering often missed temporal information, and the learned policy's effectiveness depended heavily on cluster initialization and number. More recent works have explored parameterized skills using temporal abstraction. For instance, \cite{srinivas2016option} evolved the policy gradient theorem, introducing a high-level gating policy and intra-option sub-policies. Subsequent works, such as \cite{pmlr-v70-vezhnevets17a}, proposed end-to-end training methods, alternating between high-level \textit{managers} and low-level \textit{workers}. This approach was further refined by incorporating maximum-entropy RL\cite{pertsch2021accelerating}, adversarial training\cite{lee2021adversarial}, model-based RL~\cite{shi2022skill}, meta RL~\cite{nam2022skill} and constraint-conditioned RL~\cite{liu2023constrained, yao2024constraint}.
Nevertheless, applying skill-based RL and planning frameworks onto the real robot systems is still an unsolved problem that has been marginally explored~\cite{wang2023mimicplay, xu2023xskill}, none of which addressed the contact-rich manipulation problem under tactile sensing. 


\textbf{Offline Reinforcement Learning}
Offline RL, also known as batch RL, focuses on learning policies from previously collected data without further interactions with the environment. This approach is crucial for scenarios where online data collection is expensive or risky. Conservativeness has been adopted in value and density estimation~\cite{kumar2020conservative, cen2023learning} to address the overestimation issues. TD3-BC~\cite{fujimoto2021minimalist} combines the strengths of TD3, a popular actor-critic method, with Behavioral Cloning. The Decision Transformer (DT)-based approaches~\cite{chen2021decision, liu2023constrained} rethink the RL paradigm by treating it as a sequence modeling problem, leveraging transformers to predict future rewards. Notably, DT tokenizes the reward, observation, and action from offline trajectories into different tokens, enabling the potential to incorporate multimodal inputs from the observation space. 
\begin{figure*}[h]
    \centering
    \includegraphics[width=0.75\textwidth]{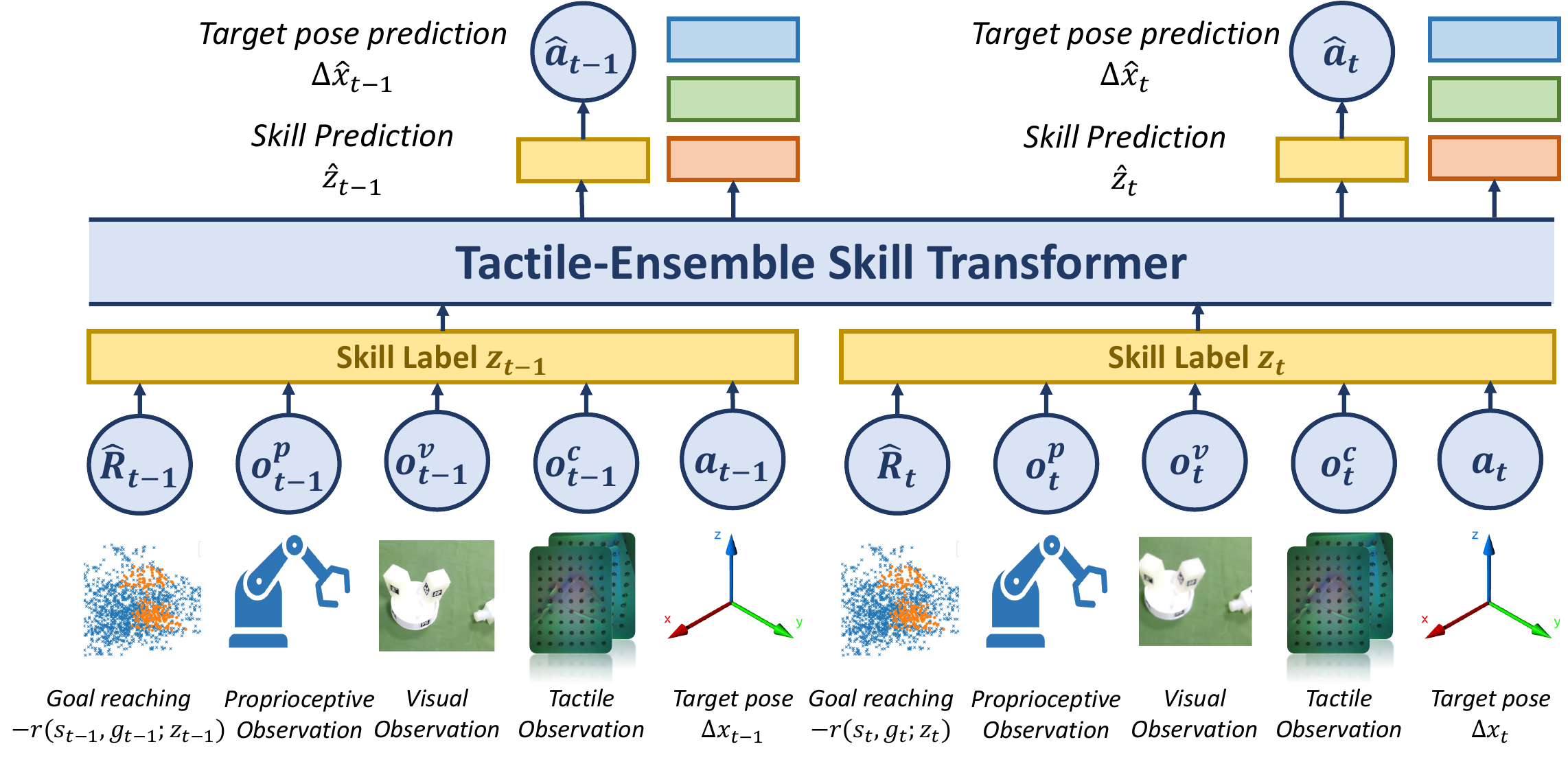}
    \caption{Diagram of TEST. We tokenize the input sequence with reward-to-go, proprioceptive observation, visual observation, followed by a tactile observations. The input bundle will predict the target pose as the action for the current timestep. At each step, the inputs are aggregated to predict the state at the current timestep. }
    \label{fig:test}
\end{figure*}
\section{Problem Formulation}
\label{sec:objective}
\textbf{Task Objective} \ The objective of TEST is to improve the quality of Learning from Imperfect Demonstration~(LfID) for long-horizon robotic assembly tasks. 
Assume we have $N$ skill primitives and a skill set denoted as $\{z^{(i)}\}_{i=1}^N$. We are given a skill-labeled offline dataset by some heuristic behavior policy $\pi_0^{(i)}$, where $(i)$ refers to the skill index of $z$. 
In general, the objective of the furniture assembly task includes two parts: \textit{accuracy} and \textit{efficiency}. For the accuracy of assembly, we evaluate the accuracy via the Average Success Rate~(ASR), i.e. $\text{ASR}=\frac{\text{\# tasks succeeded}}{\text{\# all tasks}}$, which indicates success in different assembly tasks or sub-tasks. For the efficiency of assembly, we evaluate the Average Steps~(AS), where $\text{AS}=\frac{\text{\# timesteps}}{\text{\# skill phase}}$. To better evaluate the quality of the goal-reaching quality in the learned policy, we will also consider the Average Reward~(AR) as one of the metrics.

\textbf{Framework Formulation}  \ We formulate our problem in the Goal-conditioned Partially Observable Markov Decision Process~(GC-POMDP) following the formulation of GC-MDP~\cite{hansen2022bisimulation} and POMDP~\cite{lovejoy1991survey}. 
A GC-POMDP is defined as a tuple $(\mathcal{S}, \mathcal{A}, \mathcal{O}, \mathcal{P}, \mathcal{R}, \mathcal{G}, \Omega)$, where 
$\mathcal{S}$ is the state space, here we define states as the 6D pose of the objects of interest. 
$\mathcal{A}$ is the action space that indicates the target pose and movement of the end-effector. 
$\mathcal{O}$ is a finite set of observations, and our robotic assembly system, in fact, gives us multimodal observations $\bm{c}=[o^p, o^v, o^c]$, where $o^p$ is the proprioceptive observation of the manipulator, $o^v$ represents the vision observation from an external camera, and $o^c$ refers to the contact-aware observation given by the tactile sensors. 
$\mathcal{P}$ is the state transition probability function. 
$\mathcal{G}$ is the goal space in the 6D pose of the objects to be assembled together, $G\subset S$. 
$\mathcal{R}: \mathcal{S} \times \mathcal{A} \rightarrow \mathbb{R}$ is the reward function, in practice our reward function is induced by the target goal $g\in G$. 
$\Omega: S \times A \rightarrow O$ is the observation function, which maps a state-action pair to an observation. It captures the probability of observing $o$ after taking action $a$ and ending up in state $s'$, i.e., $\Omega(o|s', a)$. 
The objective in GC-POMDP is to find a policy that maximizes the expected cumulative reward $\mathbb{E}_{\pi} \Big[\sum_{t=0}^T \gamma^t r_t|O_t\Big]$ over time.

\textbf{Additional Assumptions} \ We further model our robotic assembly task by adopting the skill learning formulation~\cite{srinivas2016option} in the above GC-POMDP. 
We represent the skill-based RL problem as a tuple $(I_z, \pi_z, \beta_z)$ associated with certain skill $z$. $I_z$ is the initial set of states of skill $z$, $\pi_z=\pi(\cdot | o, z)$ is a goal-conditioned skill-conditioned policy, and $\beta_z: \mathcal{S}\to [0, 1]$ is a termination function of the skill $z$. 

Firstly and most importantly, we assume the \textit{invariance} of skill primitives across different assembly tasks. The skill primitives required to finish the assembly tasks during testing is the superset of skills demonstrated in the training environments, i.e. $z_\text{train} \subseteq z_{\text{test}}$. 
Secondly, we assume whenever the end-effector reaches the goal of skill $z$, the manipulator always has \textit{smooth transition} to the next candidate skill in the assembly tasks, i.e. $\exists\ z'$, $\forall\  G_{z}=\{s | \beta_{z}(s)=1\}$, $G_{z}\subset I_{z'}$. 

\section{Methodology}
In this section, we introduce our proposed framework TEST. We use a Skill Transition Model~(STM), which learns the higher-level transition model $p(z'|z, \bm{c})$. 
Then for each sub-skill, we learn the intra-skill goal-reaching policies $\pi(\cdot|o, z)$ via Tactile Ensemble Policy Optimization~(TEPO), which transforms offline RL into a sequential modeling problem with hindsight relabeling as data augmentation. 
We implement both STM and TEPO in an end-to-end Tactile Ensemble Skill Transformer, which is visualized in Figure~\ref{fig:test}.
Lastly, we introduce the hierarchical skill transfer pipeline of TEST during the online deployment stage that aims the maximize the zero-shot performance in the target domain and improve robustness against sensor noise, \textit{i.e.} image corruption.

\begin{algorithm}[hbt!]
\caption{TEST Training and Inference}
\label{alg:test}
\KwData{Number of Skills $N$, number of trajectories $M$, number of iterations $K$, maximum timesteps $T$, offline behavior policy $\{\pi_0^{(i)}\}_{i=1}^N$, step-wise penalty $c$, pose distance measure $d_z$, initial State Set $I_z$, terminal state condition $\beta_z$}
\KwResult{Optimized STM $\hat{p}_\theta$, skill policies $\{\hat{\pi}_\phi^{(k)}\}_{k=1}^N$}

\tcc{TEST Hierarchical Training}
\text{Offline data collection: } $\mathcal{D}=\{\tau\}_{m=1}^M \gets \{\pi_0^{(i)}\}_{i=1}^N$\;  
\For{$k=1,\cdots, K$} {
    \text{Reward modeling: } $r(s,g;z) = -r_{\text{penalty}}-d(s,g; z)+ \alpha \mathbb{I}(s=g)$\;
    Initialize $s_0\sim I_{z}$\; 
    \tcc{Hindsight Relabeling}
    Sample goal: $g\sim p_s(\tau)$\; 
    Relabel: $\tau' = \text{Relabel}(\tau)$ with~(\ref{eq:hindsight})\;
    Data Augmentation: $\mathcal{D}\cup \{\tau'\}$\;
    \For{$m=1, \cdots, M$} {        
        \text{Sample context $\bm{c}_t'\sim \mathcal{B}_k$ with a horizon $H$}: \\
        $\bm{c}_{t'}\gets \{R_t, o_t^p, o_t^v, o_t^c, s_t, a_t\}_{t=t'-H+1}^{t'} \backslash \{a_{t'}\} $\;
        \tcc{STM}
        \text{Next skill sampling: } $z_{t'}\sim \hat{p}_\theta(\cdot | z_{t'-1}, \bm{c}_{t'})$\;
        \text{Update STM }$\theta\gets \theta - \eta \nabla_\theta \mathcal{L}_{\text{STM}}$ \ with (\ref{eq:stm})\;
        \tcc{TEPO}
        \text{Skill-conditioned policy: } $a_{t'+1} \sim \pi_\phi (a|\bm{c}_{t'}, z_{t'})$\;
        \text{Update TEPO }$\phi\gets \phi - \eta \nabla_\phi \mathcal{L}_{\text{TEPO}}$ \ with (\ref{eq:nll})\;
    }
}
\tcc{TEST Hierarchical Inference}
$z\gets z_0$\;
Initialize $s_0\sim I_{z}$\; 
\For{$t=1,\cdots, T$} {
    \tcc{Skill-based policy rollout}
    \While{!$\beta_z(s_i)$}{
        $a_{i} \gets \arg \max_{a} \pi_\phi(a|\bm{c}_{i}, z)$\;
        $o_i, r_i, \beta_z(s_i) \gets \text{env.step}(a_i; z)$\;
        $\hat{R}_{i}\gets \max(\hat{R}_{i-1}-r_i, 0)$\;
        $\bm{c}_{i}$.update(\{$\hat{R}_{i}, o_i, a_i$\})\;
        $i\gets i+1$\;
    }
    \tcc{Switch skill at termination}
    STM Prediction: $z \gets \arg \max_{z'} \hat{p}_\theta(z' | z, \bm{c}_{i})$\;
}
\end{algorithm}

\subsection{Learning the Skill Transition Model from Offline Data}
Our goal is to learn the State Transition Model as an inter-skill transition model that operates at a high level, focusing on how different skills or sub-tasks can be chained together to achieve a complex, long-horizon task. 

We have an input trajectory $\tau=\{\tau_i\}_{i=1}^T$ with a skill horizon $T$. For each $i\in [1, T]$, we have
\begin{equation}
    \label{eq:trajectory}
    \tau_i = \{o_0, a_0, r_0, s_1, \cdots, o_{T-1}, a_{T-1}, r_{T-1}; g\}
\end{equation}

The step reward in our method is goal-conditioned, labeled by the sequential information from demonstrated trajectories: 
\begin{equation}
\label{eq:reward}
r(s_t, g_t; z) = \underbrace{-r_{\text{penalty}}}_\text{Time Penalty} \underbrace{- d(s_t, g_t; z)}_\text{Distance to Goal} + \underbrace{\alpha \mathbb{I}(s_t=g_t)}_\text{Arrival Bonus},
\end{equation}

where $g_t=\{s_{t'} | \max_{t'} t'<t,\ s.t.\ \beta_z(s_{t'})=1 \}$, which is the last demonstration that satisfies the termination condition $\beta$. 
Following the autoregressive structure in~\cite{chen2021decision}, every future $z$ will depend on a context $\bm{c}$ of trajectory history, 
\begin{equation}
\bm{c}_{t}=\{R_{t-H+1}, o_{t-H+1}, a_{t-H+1}, \cdots R_t, o_t, a_t\},     
\end{equation}
where $R_{t}=\sum_{t'\geq t} r_{t'}$ is the summation of the future reward till the end of the episode, denoted as reward-to-go.  
The inter-skill transition determines the sequence in which different skills should be executed, ensuring smooth execution between consecutive trajectories of the skills. 
The STM follows a categorical distribution: $p_\theta(z'|z, \bm{c}) = \text{Categorical}(\ell_\theta(z, \bm{c}))$, 
where $\ell_\theta(\cdot, \cdot)$ is the output logits of decoder output followed by the Skill Transformer's encoder, as shown in the skill prediction block of Figure~\ref{fig:test}. 
It also considers potential dependencies between skills, ensuring that prerequisite tasks are completed before dependent ones. 

In the demonstration collection phase, we randomly sample from the heuristic policy with a Finite State Machine~(FSM). We then fit the skill transfer based on the trajectory observation $\bm{c}$ and current skill $z$ by minimizing the negative log-likelihood loss: 
\begin{equation}
    \label{eq:stm}
    \mathcal{L}_{\text{STM}} = \mathbb{E}_{\tau\sim \pi_0} \mathbb{E}_{z\sim \tau} \Big[- \log p_\theta(z'|z, \bm{c}) \Big], 
\end{equation}
By leveraging tactile feedback and ensemble learning, the inter-skill policy can make real-time decisions about what would be the most likely next skills to perform, enabling diverse way of skill composition in online deployment.

\subsection{Skill-conditioned Goal-reaching Policy Optimization}
The Tactile Ensemble Policy Optimization~(TEPO) module in the TEST framework is designed to learn a skill-conditioned goal-reaching policy $\pi(a| \bm{c}, g, z)$, where the goal is implicitly induced by $g=\{s | \beta_z(s)=1\}$. Without loss of generality, we still denote $\pi(a|\bm{c}, g, z)\triangleq \pi(a|\bm{c}, z)$. 
We parameterize our action distribution by the output logits as follows a Gaussian distribution: $\pi_\theta(a|\bm{c}, z) = \mathcal{N}\Big(\mu_\theta(\bm{c}, z), \Sigma_\theta(\bm{c}, z) \Big)$. 

Intuitively, TEPO learns a goal-reaching policy at the sub-skill level. Although the horizon is significantly shortened compared to directly learning over the entire horizon of tasks, the rewards could still be sparse, being provided only when the exact goal is achieved. 
This sparsity can adversely affect learning, especially in our offline settings where the robot cannot interact with the environment to gather more data. 
Therefore, we conduct an additional goal relabeling strategy for TEPO training. 
For the input sub-skill trajectory $\tau_k$ corresponding to $z_k$ introduced in~(\ref{eq:trajectory}), the original $g\in \{s | \beta_{z_k}(s)=1\}$. 
We then resample the goal states from those in trajectories $\tau_k$,  
\begin{equation}
\label{eq:hindsight}
\begin{aligned}
    & \text{Goal Relabeling: } g'\sim p_s(\tau_k) \\
    & \text{Reward Relabeling: } r_t' = r(s, g'; z_k),
\end{aligned}
\end{equation}
where $p_s(\cdot)$ is the empirical marginal state distribution of the input trajectories. 
After the hindsight relabeling, we can generate multiple relabeled trajectories $\tau_k' = \{o_0, a_0, r_0', s_1, \cdots, o_{T-1}, a_{T-1}, r_{T-1}'; g'\}$, which diversifies the step reward, and corresponding reward-to-go predictions for identical historical sequences, improving goal scenario generalization.
After the data augmentation with hindsight relabeling, we get the augmented trajectories $s$. 
Given the offline demonstration, TEPO aims to minimize the following negative log-likelihood loss with an entropy regularizer~\cite{haarnoja2018soft}: 
\begin{equation}
    \label{eq:nll}
    \mathcal{L}_{\text{TEPO}} = \mathbb{E}_{\tau_z\sim \pi_0^{z}} \Big[ -\log \pi_\phi(a | \bm{c}, z) - \lambda H[\pi_\phi(\cdot | \bm{c}, z)] \Big].
\end{equation}
where $\lambda$ is the weight of the regularizer. 
The learned low-level policy $\pi(a|\bm{c}, z)$ takes into account both the context $\bm{c}$ consisting of multimodal observations, goal preference, and skill representation $z$. Through a combination of tactile feedback and ensemble learning, the intra-skill policy optimizes the trajectory in real-time, ensuring that the robot can adapt to changes and uncertainties in the environment. This adaptability is crucial for tasks like \textit{insertion} that require fine motor skills, such as aligning parts with tight tolerances.
The training pipeline is summarized as a pseudocode in Algorithm~\ref{alg:test}. After we train the TEST model with STM and TEPO in an alternative optimization, we apply hierarchical inference at the online deployment stage to further improve the performance of TEST. 
As illustrated in Algorithm~\ref{alg:test}, TEST conducts hierarchical inference between the skill-conditioned goal-reaching policies and skill transition predictions. 
TEST follows the transformer structure of GPT-2~\cite{radford2019language}. 


\section{Experiments}

\subsection{Environment Design} 
\textbf{Tasks Design:} We design our furniture assembly platform based on the FurnitureBench set~\cite{heo2023furniturebench}. Specifically, we select the one-leg assembly of the square table, the most widely studied environment as our in-distribution setting. For the generalization setting, we selected a different furniture stool with a different leg geometry and different angles of threading. Similar to the in-distribution setting, here we only consider the one-leg assembly task for real-robot evaluation. 

\begin{figure}[h]
    \centering
    \includegraphics[width=0.40\textwidth]{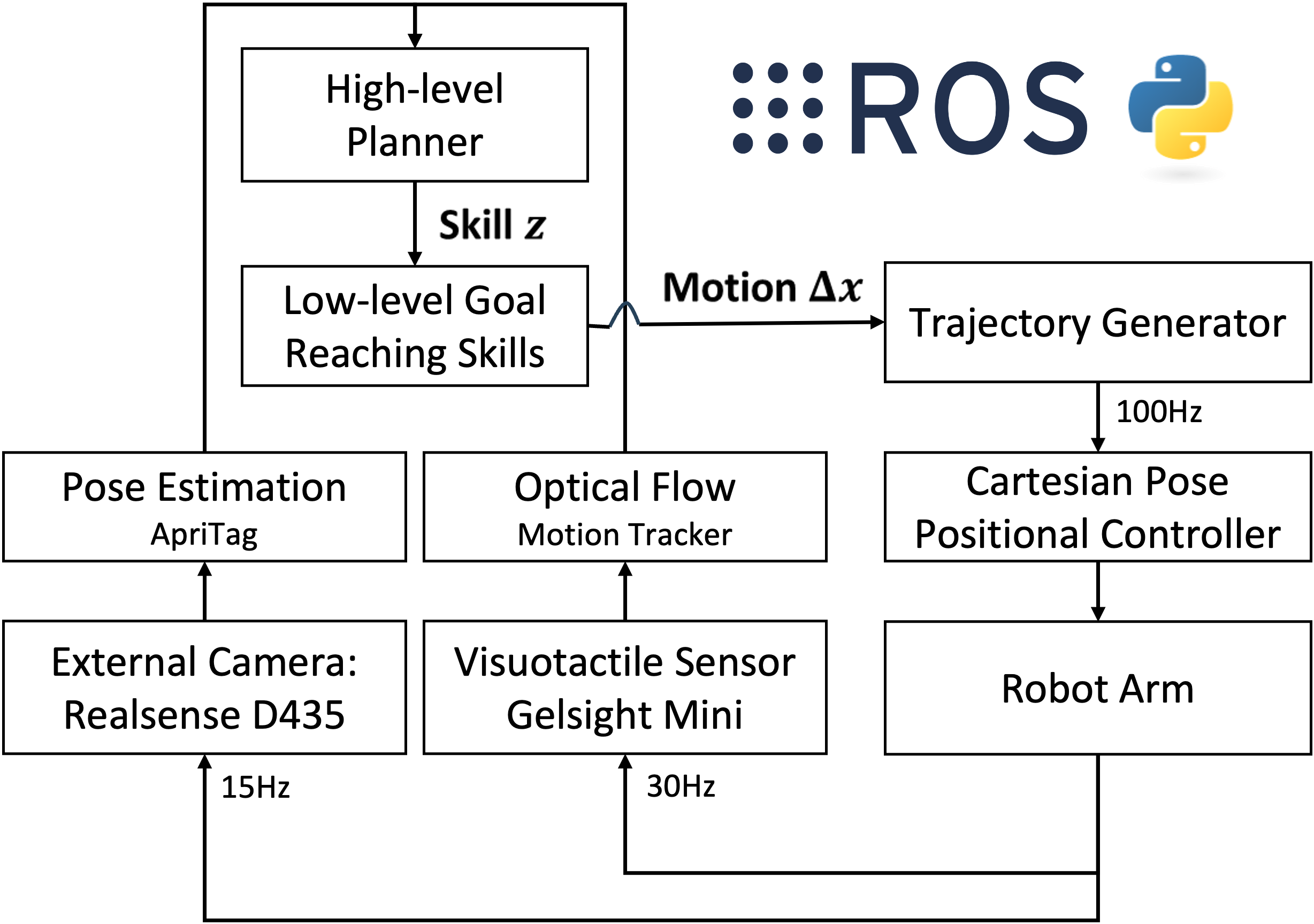}
    \caption{Hardware system setup for the experiments. }
    \label{fig:hardware}
\end{figure}

\textbf{Hardware Setup:} As is illustrated in Figure~\ref{fig:hardware}, to support our hierarchical decision-making systems with multimodal sensory inputs, we use a Kinova Gen3 collaborative robot arm as the manipulator to which we instrumented its gripper fingers with tactile sensing devices. We chose the Gelsight Mini~\cite{yuan2017gelsight}, a type of optical-based tactile sensor with excellent optical resolving power. We are also using the optional tracker gel pads which give us additional feedback about the slipping state of the grasped parts. The sampling frequency is 15 Hz on the external camera and 30 Hz on the tactile sensors on both fingers. In the low-level control, we use position control in the end-effector's Cartesian space at 100 Hz which allows us to generate smooth interpolated trajectories. 

\textbf{Real-world Offline Data Collection: } In the real-world experiments, we use the heuristic policies $\pi_0^{(i)}$, where $z^{(i)}\in\{\text{pick, reach, insert, screw, adjust}\}$. The skill is parameterized by the starting pose and goal 6D pose of the end effector. 
During the data collection phase, we use AprilTags to represent the objects' state, then use the estimated state to design goal-reaching policies with some randomness. To guarantee the safety of the real robot, we actively detect the violation of safe contact constraints by the movement of the tactile sensor's markers. Specifically, we use the optical flow to detect such violations. We collected a total of 2,000 trajectories for all the skills and this heuristic policy is fully operated in the real world. 

\vspace{-0.3cm}

\begin{table}[hbt!]
\small{
    \centering
    \begin{tabular}{c | c c c} \bottomrule
    \multirow{2}{*}{\textbf{Method}} & \multicolumn{3}{c}{\textbf{Square Table} / \textbf{Stool}}  \\
    & AR~($\uparrow$)& ASR~($\uparrow$) & AS~($\downarrow$) \\ \midrule
    BC+GSA & 0.59 / 0.14 & 0.3 / 0.0 & 36.8 / 80.0\\
    LSTM+GMM & 1.02 / 1.05 & 0.8 / 0.5 & 22.4 / 24.4 \\ \midrule
    TEST w/o STM & 0.33 / 0.16 & 0.4 / 0.2 & 49.0 / 76.4 \\
    TEST w/o TEPO & 0.85 / 1.12 & 0.7 / 0.4 & 62.0 / 66.4 \\
    TEST~(Ours) & \textbf{1.64} / \textbf{1.52} & \textbf{0.9} / \textbf{0.9} & \textbf{16.4} / \textbf{14.0} \\ \midrule
    Heuristic Policy & 1.00 / 1.00 & 0.7 / 0.7 & 67.1 / 64.0 \\ \bottomrule
    \end{tabular} 
    \caption{Results on online assembly of square table, and generalization setting for the stool. The evaluation results average over 10 episodes on our furniture assembly robot. The \textbf{Bold} means the best results among all. }
    \label{tab:generalization}
}
\end{table}
\vspace{-0.3cm}

\begin{table}[hbt!]
\small{
    \centering
    \begin{tabular}{c | c c c } \bottomrule
    \multirow{2}{*}{\textbf{Noise Level} (cm)} & \multicolumn{3}{c}{\textbf{Evaluation Metrics}} \\
    & Reward~($\uparrow$)& ASR~($\uparrow$) & AS~($\downarrow$)  \\ \midrule
    0.1 & 1.58 & 0.9 & 28.8  \\
    0.2 & 1.34 & 0.9 & 41.2 \\
    0.5 & 1.09 & 0.8 & 64.4\\
    1.0 & 0.68 & 0.5 & 77.2 \\ \midrule
    0.0 & \textbf{1.64} & \textbf{0.9} & \textbf{16.4} \\ \bottomrule
    \end{tabular} 
    \caption{Experiment results on TEST's robustness under disturbances. We add Gaussian noise to the state prediction from vision inputs. The noise level indicates the standard deviation of the target position~(on the x, y, and z-axis). }
    \label{tab:robustness}
}
\end{table}
\vspace{-0.3cm}

\subsection{Baselines}
We compare the performance of TEST with the following baselines and the variants of TEST. 
\textbf{BC+GSA}~\cite{akrour2018regularizing}, or Behavior Cloning with Generalized State Abstraction applies unsupervised clustering for state abstraction and hierarchical policy learning based on the input offline data in the entire task horizon.
\textbf{LSTM+GMM}~\cite{pastor2020bayesian}, uses Long Short-term Memory and Gaussian Mixture Model, aiming to capture both the sequential nature of the demonstrations and variability in the actions across the states. It's particularly designed for long-horizon LfD problems with multimodal observations. 
\textbf{TEST w/o STM}, is an ablation variant on TEST by removing the skill transition model. 
\textbf{TEST w/o TEPO} is an ablation variant on TEST by removing the tactile ensemble in policy optimization. 
\textbf{Heuristic Policy}: a set of heuristic policies $\{\pi_0^{(i)}\}_{i=1}^N$ that are used to collect data. Safe yet conservative, the heuristic policies have access to privileged information on the skill label during the evaluation stage.

\subsection{Evaluation Protocol}
As mentioned in Section~\ref{sec:objective}, we compare the following metrics: AR, ASR, and AS. The original definition of the reward is given by~(\ref{eq:reward}). 
Based on our furniture assembly robot, as visualized in Figure~\ref{fig:front_figure}, we evaluate all the experiments with an average of 10 episodes. The maximum number of skills per episode is 80, and we count each skill as a step in the AS metric. For the AR, we normalize the average return with respect to the heuristic policy $\pi_0$. 

\subsection{Results and Analysis}
In this part, we answer the following research questions: 
\begin{itemize}
    \item \qa{\textbf{RQ1}}: What is the in-distribution performance of TEST compared to the heuristic policies and other baselines? 
    \item \qa{\textbf{RQ2}}: What is TEST's generalization performance towards unseen furniture, compared to other baselines?
    \item \qa{\textbf{RQ3}}: How is the robustness of TEST under noise disturbances in the observation space, e.g. image corruption? 
\end{itemize}

\begin{figure}[h]
    \centering
    \includegraphics[width=0.50\textwidth]{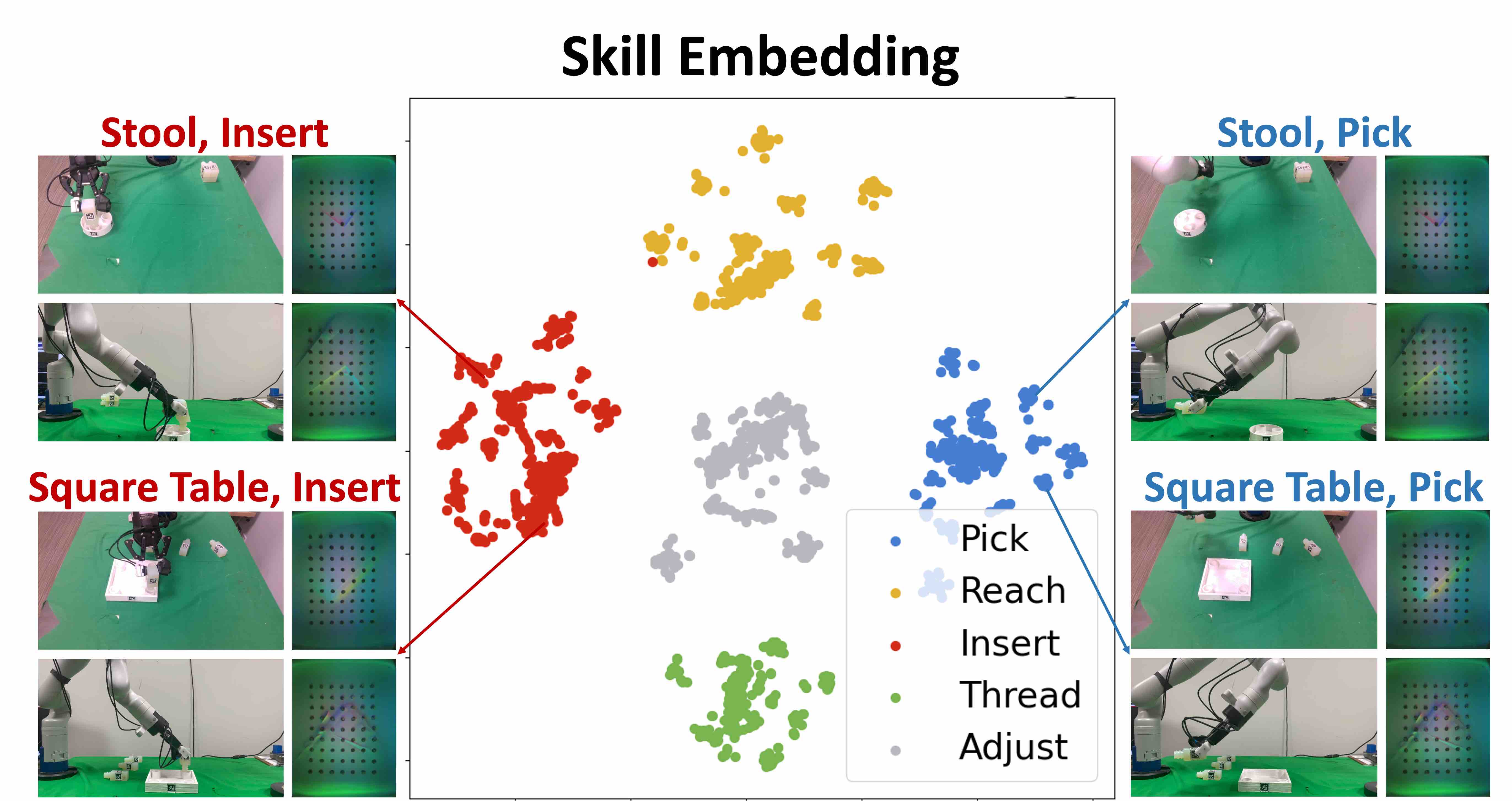}
    \caption{Visualization of the embedding of TEST's learned skills t-SNE. We see a clear cluster in each sub-skill, and rollout trajectories in the stool and square table align well. }
    \label{fig:embedding}
\end{figure}

\begin{figure}[h]
    \includegraphics[width=0.45\textwidth,left]{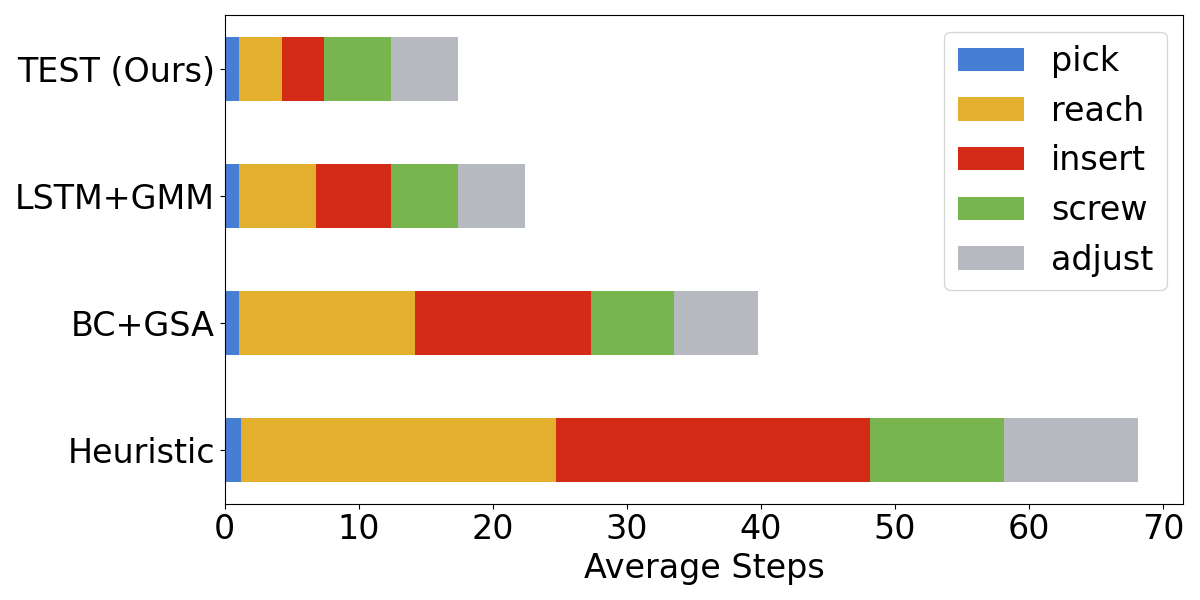}
    \caption{Bar plot for comparison on the Average Steps~($\downarrow$) of each sub-skill between four methods. We can see that TEST significantly outperforms other baselines and the heuristic behavior policy $\pi_0$. }
    \label{fig:barplot}
\end{figure}

We illustrate our key findings on the above three research questions in Table~\ref{tab:generalization},~\ref{tab:robustness} and Figure~\ref{fig:embedding},~\ref{fig:barplot}. 

For \qa{\textbf{RQ1}}, TEST outperforms BC+GSA and LSTM+GMM with higher accuracy and efficiency in the long-horizon assembly tasks in the square table. 
For \qa{\textbf{RQ2}}, TEST still manages to generalize and outperform the Heuristic Policy even if it does not access the direct skill labels in the unseen furniture stool. Compared to other baselines, TEST has the lowest performance drop and the highest success rate compared to the other baselines. 
For \qa{\textbf{RQ3}}, we manually add the Gaussian noise from the prior pose estimation from vision observation $o^v$ and evaluate TEST's performance in the square table environment. Though the efficiency of TEST drops significantly, the accuracy is still maintained to some extent. Attributed to the multi-modal design, TEST is actually robust under mild disturbances in the visual observation. 

We also provide ablation studies by removing key modules in TEST. TEST w/o STM only conducts the tactile ensemble in policy optimization for the long task horizon, which indicates the agent can hardly generalize even with contact awareness, if the skill transition is not properly represented. TEST w/o TEPO keeps the hierarchical skill-based learning structure while removing the tactile signals in policy optimization. 
This reinforces the idea that the addition of tactile sensors improves the performance of contact-rich manipulation. 

In addition, to understand the effectiveness of TEST's design, we scatter the embedding of skill representation in Figure~\ref{fig:embedding} to verify the invariance of skills between different furniture configurations. We also analyze the efficiency in each sub-skill in Figure~\ref{fig:barplot}. The results show that \textit{pick} is the easiest skill, while \textit{insertion} is the hardest one where TEST outperforms baselines with the clearest margin.  

\section{Conclusion}
In this work, we introduced TEST, a hierarchical, skill-based offline reinforcement learning framework tailored for robotic assembly tasks. TEST emphasizes the integration of tactile feedback, enhancing the contact-aware decision-making of robotic agents. At its core, TEST employs a Skill Transition Model, parameterized by a trajectory-level transformer for inter-skill transitions, and leverages Hindsight goal relabeling for intra-skill policy learning. Comprehensive evaluations on a furniture assembly robot underscore TEST's superiority over existing LfID baselines.

The limitation of TEST is that it still assumes access to accurate skill labels in offline data, which may not always be available in more general teleoperation cases. The assumption that skill primitives seamlessly integrate is another simplification, overlooking potential mismatches between consecutive skill states. Additionally, as furniture part geometries become complicated, the challenge of effectively fusing tactile imprints with marker movements for decision-making emerges as a promising direction for future research.









\newpage
\bibliographystyle{IEEEtran}
\bibliography{reference}

\end{document}